\documentclass[conference]{IEEEtran}
\IEEEoverridecommandlockouts
\usepackage{cite}
\usepackage{amsmath,amssymb,amsfonts}
\usepackage{algorithmic}
\usepackage{graphicx}
\usepackage{subcaption}
\usepackage{textcomp}
\usepackage{xcolor}
\usepackage{stfloats}
\usepackage{float}
\usepackage{hyperref}
\usepackage{etoolbox}

\makeatother

\def\BibTeX{{\rm B\kern-.05em{\sc i\kern-.025em b}\kern-.08em
    T\kern-.1667em\lower.7ex\hbox{E}\kern-.125emX}}

\begin{document}

\title{Virtual Ring Try-On}

\author{\IEEEauthorblockN{Vishnu D. Burkhawala}
\IEEEauthorblockA{\textit{Department of Information Technology} \\
\textit{Dharmsinh Desai University}\\
Nadiad, India \\
burkhawalavishnu@gmail.com \\
}
\and 
\IEEEauthorblockN{Zankhana J. Barad}
\IEEEauthorblockA{\textit{Department of Information Technology} \\
\textit{Dharmsinh Desai University}\\
Nadiad, India \\
zankhana.it@ddu.ac.in \\}
\and
\IEEEauthorblockN{Harshadkumar B. Prajapati}
\IEEEauthorblockA{\textit{Department of Information Technology} \\
\textit{Dharmsinh Desai University}\\
Nadiad, India \\
prajapatihb.it@ddu.ac.in \\}
\and 
\IEEEauthorblockN{Vipul K. Dabhi}
\IEEEauthorblockA{\textit{Department of Information Technology} \\
\textit{Dharmsinh Desai University}\\
Nadiad, India \\
vipuldabhi.it@ddu.ac.in \\}
}

\maketitle

\begin{abstract}
This paper presents an innovative approach that enables the users to capture their hand and try the jewel ring on their hand. The user captures the image of the hand using the React Native base GUI of the mobile application and selects the ring that the user wants to try, and the output image will have the user's hand with the ring image. This approach is implemented using a combination of MediaPipe [5] hand point detection and YOLO-V8 [6] custom object detection. The hand image uploaded by the user first undergoes mediapipe [5] hand point detection. It will give the hand points and a Region of Interest mask where the ring is going to be placed. Then the ring is passed through YOLO [6] object detection, in which ring points are detected, and background is removed. After that, using vector algebra, the angular discrepancy between the finger’s reference axis and the ring’s principal axis is computed. Also, ring size is rescaled according to finger thickness, preserving the aspect ratio to maintain perceptual realism. Then the ring is placed on the hand image and the output image is generated and shown on the user screen.
\end{abstract}


\section{Introduction}

In today's era, where online shopping is widely accepted, the jewelry market embraces virtual try-on features to increase the customer experience and purchase certainty. A system that includes realistic virtual try-ons allows users to see how a product will look in their hand before purchasing the product. Such a system helps to eliminate doubts in the customer's mind and increase the conversion rate of the website and online stores.

Many approaches have been put out in recent years to deal with this issue. Examples of model-based picture generation systems that try to generate fresh images by simultaneously regenerating the hand and ring are generative models and large-scale diffusion models. These methods are effective for creating high-quality images in some cases, but they need computationally powerful GPUs, large-scale models that have already been trained, and exact text/image instructions to yield excellent results. Furthermore, the worldwide regeneration procedure tends to alter the hand's look, compromising authenticity. In contrast, the suggested system uses no deep generative algorithms, requires no GPU or other specialized hardware, and runs smoothly using lightweight geometric computation driven by a single user input.

Another popular class of solutions is augmented reality (AR)-based 3D modeling. These methods use hand posture predictions to spin jewelry dynamically and convert it into 3D representations. Though potentially immersive, these systems are often hardware-dependent, and the only devices that can render smoothly are costly mobile devices or AR-enabled headgear.  On mid-range devices, output is frequently laggy, jittery, or misaligned, so it does not look so realistic. Also, with this approach, the ring looks like it is floating on the finger, not fixed and attached, because it can not handle occlusion and shadows. The proposed system can handle this and gives output where the ring is fixed on the finger, not floating, and also, this system does not require high-end devices, ensuring accurate geometric alignment for jewel attachment.

Finally, 2D overlay-based approaches constitute the simplest solution. These methods generally use Mediapipe [5] to detect hand landmarks and then place a static image of the ring at the location on the finger. However, such type of overlay does not adapt to the rotation of the finger, lighting and shadows, thus looking very unrealistic, especially when the hand is side-facing. In contrast, the proposed system uses angle-aware adjustments so that the ring gets aligned with the hand's perspective, thereby maintaining realism across varying poses.

In summary, current methods either depend on computationally and GPU-required expensive generative pipelines, hardware-intensive AR rendering, or oversimplified static overlays. The proposed system is based on lightweight geometric computation and achieves a balance between realism and efficiency. By preserving the natural look of the hand while ensuring adaptive and natural-looking ring placement, this work contributes to a newer direction in the design of an accessible, realistic, and user-friendly virtual jewelry try-on system.

\section{Related Work}

Various approaches have been suggested in the field of virtual try-ons. Each method is based on different computational ideology, from classical overlays to state-of-the-art generative models. Although these approaches have made possible significant advances, they are limited by some constraints, such as limited realism, usability, and suitability for deployment in real-world consumer environments.

\subsection{Overlay-based approaches}

Overlay-based approaches represent the simplest implementation of jewelry try-on. As discussed in [1], these systems primarily employ landmark detection techniques such as MediaPipe [5] to localize hand regions and statically place a 2D ring image on the detected finger. These techniques are computationally lightweight and not so complex, but they fail to incorporate finger orientation or depth estimation. Hence, when the hand is rotated or viewed from different angles, the ring still remains fixed in a front-facing projection. That is why it leads to an unnatural and awkward appearance. This system lacks dynamic adaptation. Therefore, the realism is significantly reduced. So, for these reasons, the applicability of these systems on real-world e-commerce platforms is restricted where customers expect a lifelike experience.

\subsection{Augmented Reality (AR)-based solutions}

Augmented Reality (AR)-based solutions attempt to address this gap by employing 3D models of jewelry items that rotate in synchronization with hand pose estimation. For instance, the work presented in [2] explores AR-based try-on pipelines. While this technique theoretically improves immersion, it is highly dependent on device capabilities. For smooth rendering, it requires high-end AR-enabled smartphones, and on mid-range devices, the outputs usually exhibit lag, jitter, and reduced resolution. Also, noise in pose tracking results in misalignments that cause the ring to slide or appear floating on the finger, not fixed or attached to the finger surface. Such artifacts significantly reduce realism, and the demands of high computational devices limit accessibility for widespread consumer use.

\subsection{Generative image synthesis approaches}

Generative image synthesis approaches, such as SDEdit [3], explore image regeneration techniques that can actually redraw parts or regions of an image based on stochastic differential equations. These methods inherently perform global or semi-global image regeneration, wherein the model attempts to resynthesize both the masked region and surrounding context. While powerful, such generative models require massive GPU resources, extensive training datasets, and precise prompt engineering. More critically, due to their stochastic nature, they may inadvertently alter unmasked regions of the image—such as the background or even the appearance of the user’s hand. In jewelry try-on, this is undesirable, as users prefer that the original hand not be changed except that the ring is superimposed. Therefore, although generative pipelines exhibit robust power in artistic synthesis, computational weight and the absence of pixel-level strict control limit their application toward this purpose.

\subsection{Diffusion-based localized inpainting methods}

A variation of generative editing is seen in diffusion-based local inpainting techniques, such as SmartBrush [4]. In these, text and shape guidance are utilized to restrict the generative process to a masked area, allowing for local object replacement without regenerating the entire scene globally. This process is less intrusive than full-image synthesis but still computationally intensive, needing GPUs and diffusion pipelines that are heavy even when performing localized edits. Additionally, as in [4], inpainting using diffusion could not perfectly replicate the provided reference object. In jewelry try-on applications, this takes the form of an inability to produce the same ring image the user has given, tending to create minor variations or simply different designs. The stochastic process of diffusion thus sacrifices fidelity, and dependence on heavy computation resources restricts implementations on consumer-level devices.

In summary, overlay-based techniques are unrealistic, AR-based techniques require high-end hardware but remain artificial-looking, and generative techniques are limited by computational complexity and replicability issues. These constraints drive the necessity for a light yet realistic technique like the one presented in this work that integrates geometric alignment with direct image preservation to obtain high visual quality without relying on GPU.

\section{Proposed System}
The proposed system uses computer vision and geometric transformation to offer a realistic jewelry try-on experience without relying on large-scale AI frameworks or GPU computation. The framework consists of four primary steps: (i) landmark detection of hands, (ii) ring object localization, (iii) geometric alignment and resizing, and (iv) image synthesis and rendering. The above pipeline is to be executed seamlessly on mid-end smartphones with low computational costs.

\subsection{Hand Landmark Detection}
The first stage involves detecting the user's hand landmarks using MediaPipe [5]. Given an input image $I_{hand}$, MediaPipe [5] outputs a set of 21 landmark points:
\[
L_{hand} = \{ (x_i, y_i) \mid i = 1, 2, \dots, 21 \}
\]
where $(x_i, y_i)$ are the pixel coordinates of the $i^{th}$ landmark. Region of Interest (RoI) in order to place the ring is defined based on the specific landmarks according to the target finger.

\begin{figure*}[h!]
    \centering
    \includegraphics[width=1\linewidth,height=5.5cm]{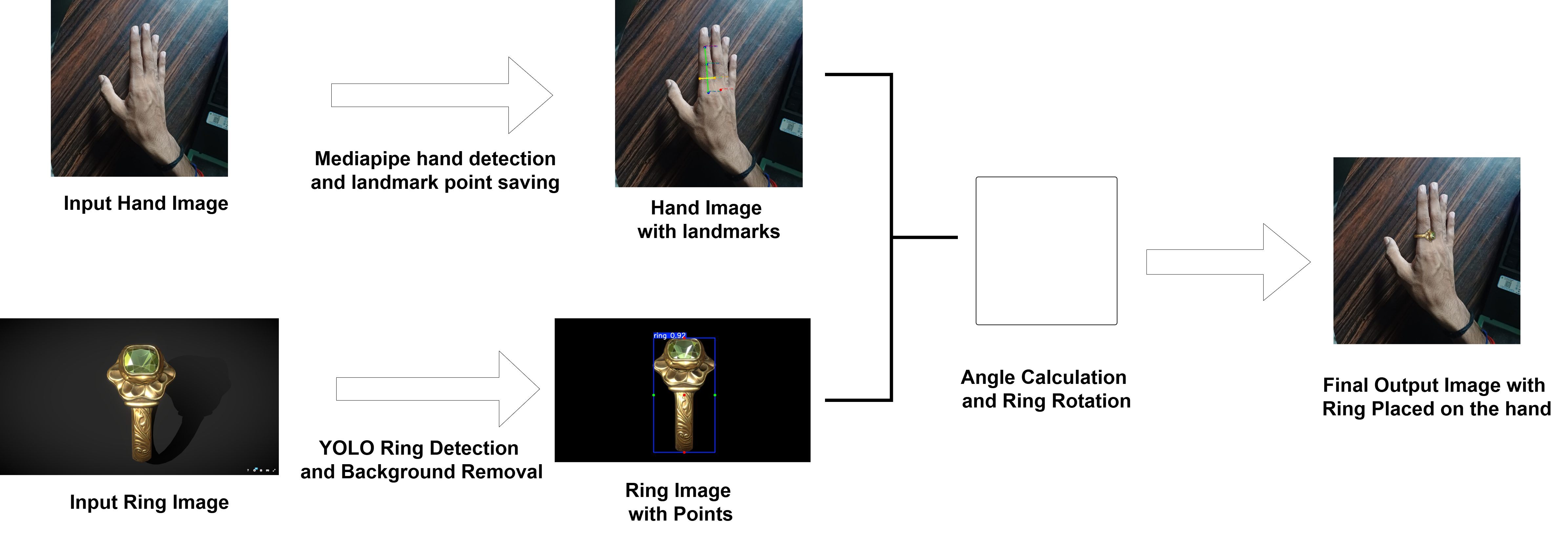}
    \caption{Workflow of proposed system}
    \label{fig:placeholder}
\end{figure*}

\subsection{Ring Object Localization}
I have fine-tuned a YOLOv8-based [6] custom object detection model for Ring detection. To train the model, I created a ring data set containing around 600 to 700 ring images with different orientations, styles, and colors. The ring image $I_{ring}$ is passed through background removal and a YOLOv8-based [6] customized object detector. The YOLO [6] model creates the bounding box around the ring that has been detected.
\[
B_{ring} = (x_{min}, y_{min}, x_{max}, y_{max})
\]
It will be used to crop and split the ring. The ring pixels will be retained while removing the unwanted background.

\subsection{Geometric Alignment and Scaling}
Once the finger RoI and ring object have been determined,
vector algebra is used to orient the ring along the finger orientation.
Let $\vec{f}$ be the reference vector of the finger axis (obtained from two hand landmarks, the base and tip of the finger), and let $\vec{r}$ be the principal axis of the ring. The angular difference is calculated as:
\[
\theta = \cos^{-1}\left( \frac{\vec{f} \cdot \vec{r}}{\|\vec{f}\| \|\vec{r}\|} \right)
\]
The ring is rotated by $\theta$ in order to be aligned with the finger's orientation. In addition, the ring is scaled by the finger width $w_f$ and the ring width $w_r$:
\[
s = \frac{w_f}{w_r}
\]
where $s$ is the scaling factor for uniform scaling to preserve
the ring's aspect ratio and thus perceive it as real.

\subsection{Image Synthesis and Rendering}
The ring image $I_{ring}^\prime$ is superimposed on the hand RoI after alignment and scaling of the ring using alpha blending:
\[
I_{output}(x, y) = \alpha \cdot I_{ring}^\prime(x, y) + (1 - \alpha) \cdot I_{hand}(x, y)
\]
where $\alpha$ represents the transparency mask. Then the output image Ioutput is displayed and presented on the user's screen.

\subsection{System Workflow}
The workflow of the proposed system is as follows :
\begin{enumerate}
    \item The user captures the hand image using React Native interface.
    \item Hand landmarks are detected using MediaPipe [5].
    \item Ring is extracted using YOLOv8 [6] custom object detection.
    \item The orientation of the ring is aligned with finger axis using vector algebra
    \item Scale the size of the ring according to the finger thickness.
    \item Overlay and render the try-on image as the final output.
\end{enumerate}

With MediaPipe [5] for accurate landmark detection and YOLOv8 [6] for object isolation, accuracy is assured, and lean geometric warping ensures efficiency. This system, unlike model-based generative methods, does not need GPU capabilities, heavy computation, or text inputs. The output is a realistic real-time jewelry try-on system deployable on hardware.

\section{Experiments and Results}

To evaluate the effectiveness of our proposed system, we conducted a comparative analysis against existing jewelry try-on solutions provided by industry platforms and research frameworks. Specifically, we tested and documented outputs from Google AI Studio (Banana Pro), OpenAI (ChatGPT) generative models, and the commercial e-commerce platform Candere by Kalyan, alongside the outputs generated by our system in fig. 2. For consistency, same input was used for all platforms to ensure a fair evaluation.
To provide a clearer visual comparison, a representative output from each model is presented in fig. 3. These examples highlight the differences in quality between the results generated by existing systems and our proposed system.

The resolution of the input ring and hand is 1024 x 1024, while the size of the ring image and the size of the hand image are in the range of 400 KB to 600 KB. I have taken photographs of real rings and also gathered images from "Sketchfab" and "Free3D" in order to have different types of ring images for experiments. To get the result from different existing systems like Google AI Studio (Banana Pro), OpenAI (ChatGPT), and Candere by Kalyan, I have used their official website.\\

\clearpage
\begin{figure*}[t!]
    \centering
    \includegraphics[width=0.85\textwidth]{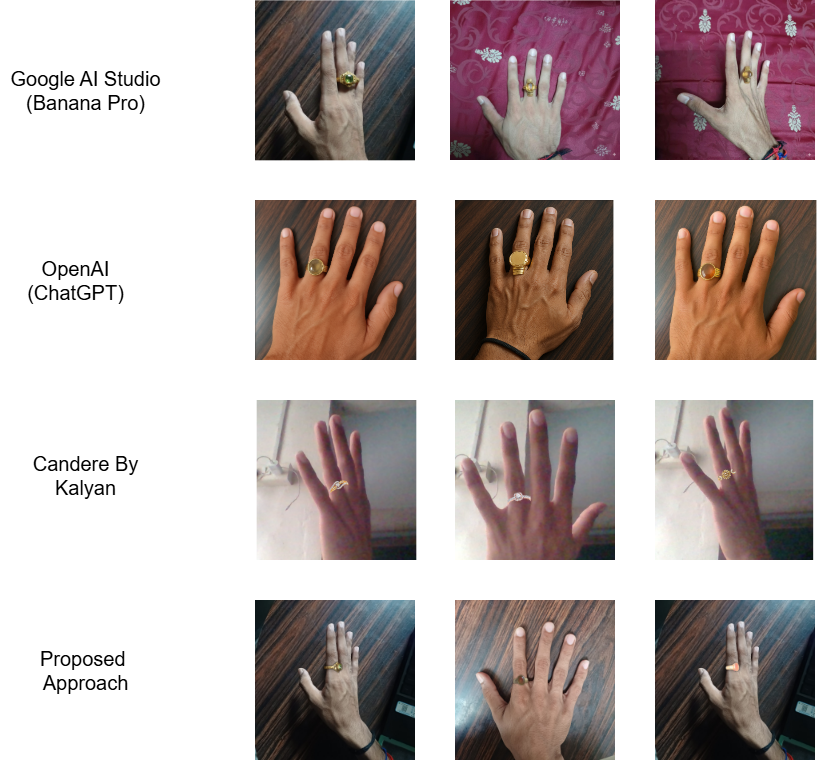}
    \caption{Comparison of outputs generated by different models and proposed approach}
    \label{fig:outputcomp}

    \hspace{2cm}
    \begin{subfigure}[b]{0.23\linewidth}
        \centering
        \includegraphics[width=\linewidth]{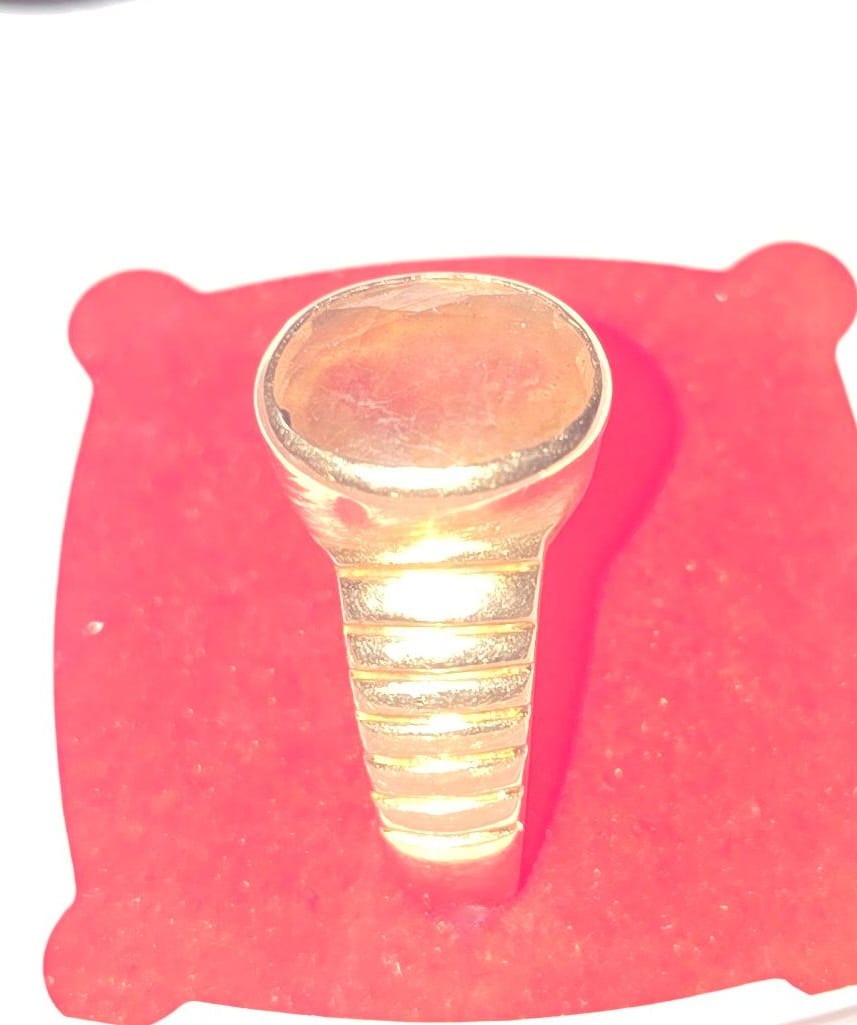}
        \caption{Input Ring}
        \label{fig:input}
    \end{subfigure}
    \hspace{3.5cm}
    \begin{subfigure}[b]{0.23\linewidth}
        \centering
        \includegraphics[width=\linewidth]{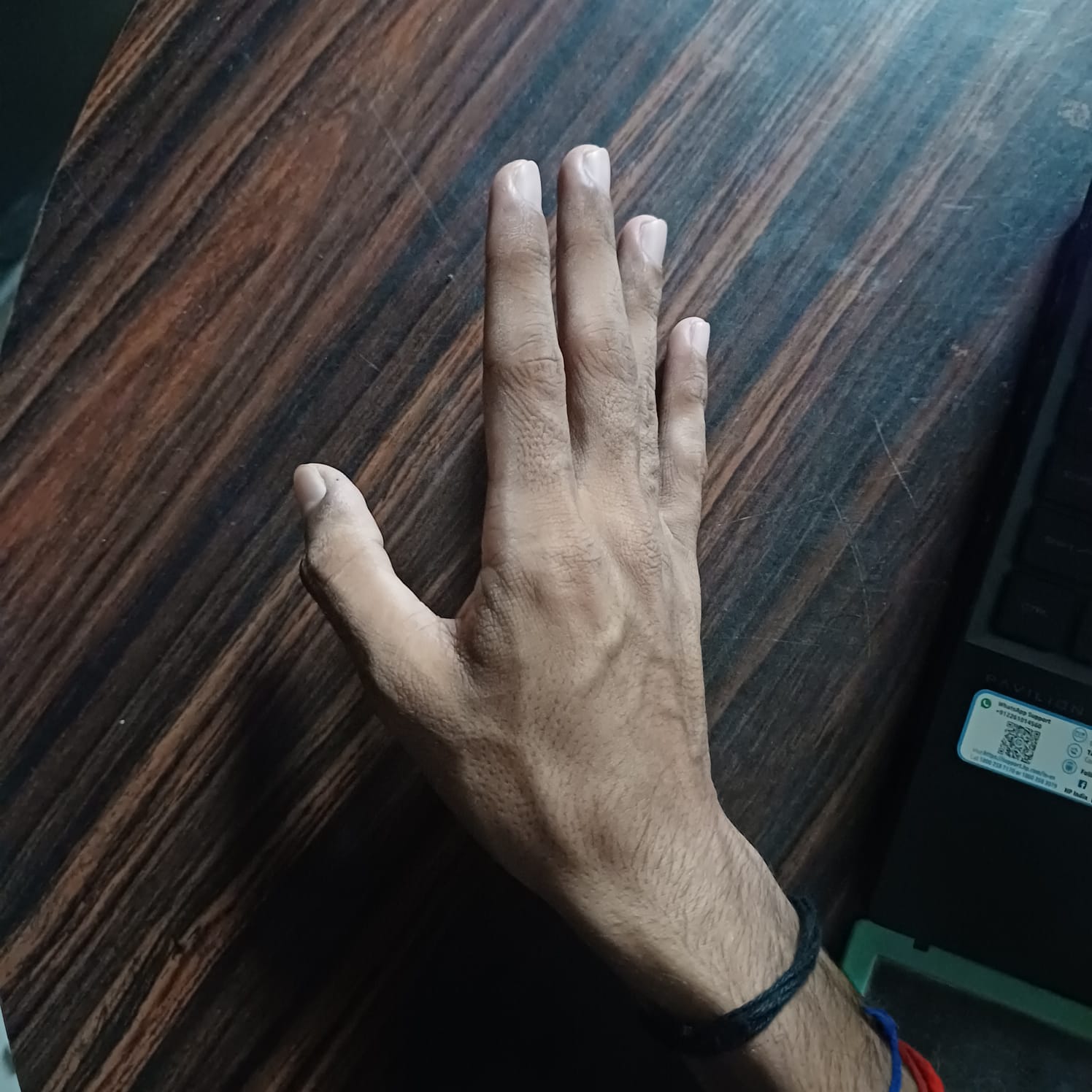}
        \caption{Input Hand}
        \label{fig:input}
    \end{subfigure}
    \newline
    \hfill
    \begin{subfigure}[b]{0.23\textwidth}
        \centering
        \includegraphics[width=\textwidth]{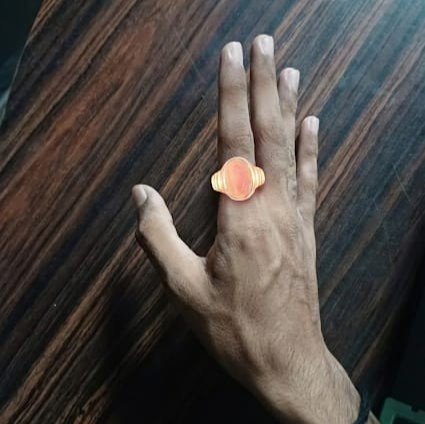}
        \caption{Google AI Studio (banana pro)}
        \label{fig:google}
    \end{subfigure}
    \hfill
    \begin{subfigure}[b]{0.23\textwidth}
        \centering
        \includegraphics[width=\textwidth]{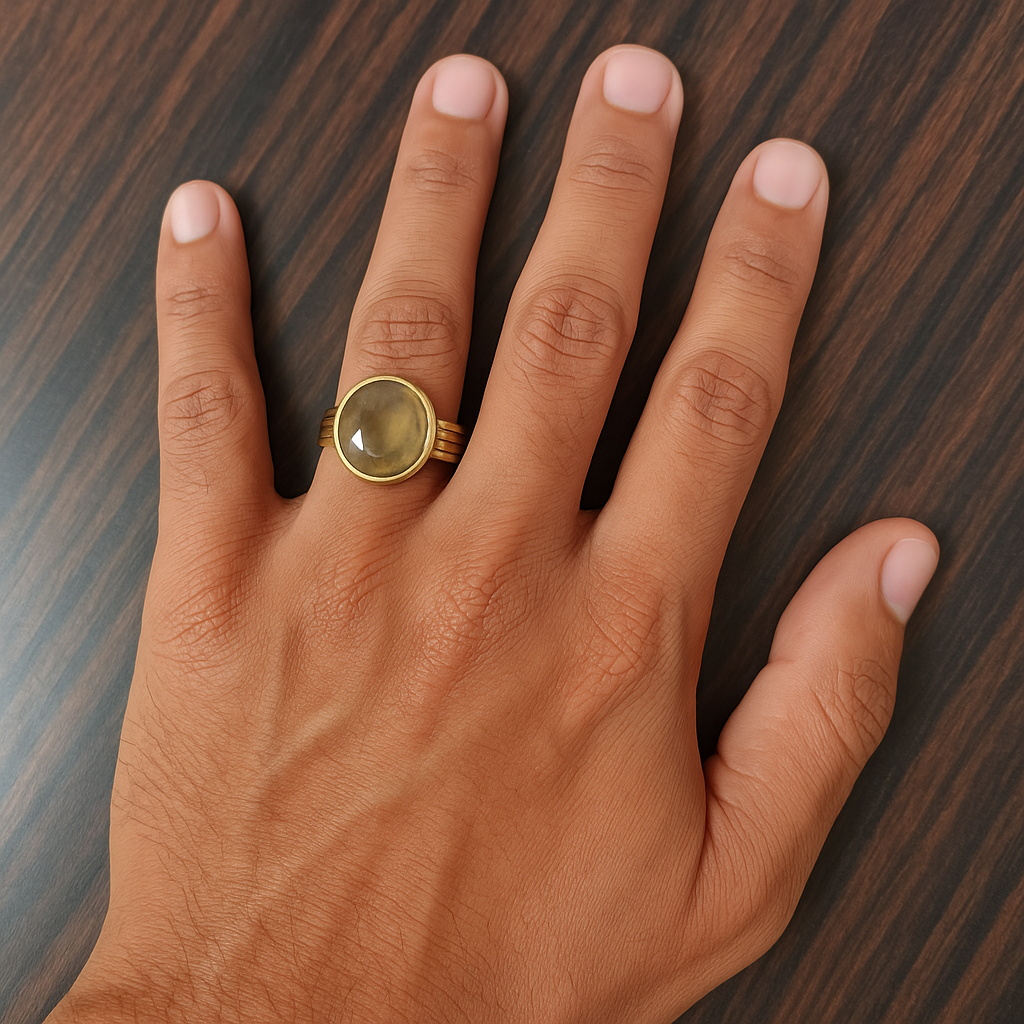}
        \caption{OpenAI (ChatGPT)}
        \label{fig:chatgpt}
    \end{subfigure}
    \hfill
    \begin{subfigure}[b]{0.23\textwidth}
        \centering
        \includegraphics[width=\textwidth]{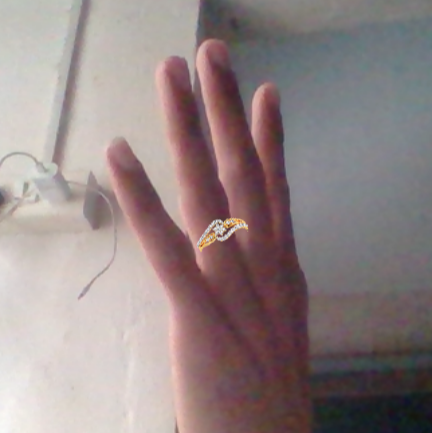}
        \caption{Candere By Kalyan}
        \label{fig:candere}
    \end{subfigure}
    \hfill
    \begin{subfigure}[b]{0.23\textwidth}
        \centering
        \includegraphics[width=\textwidth]{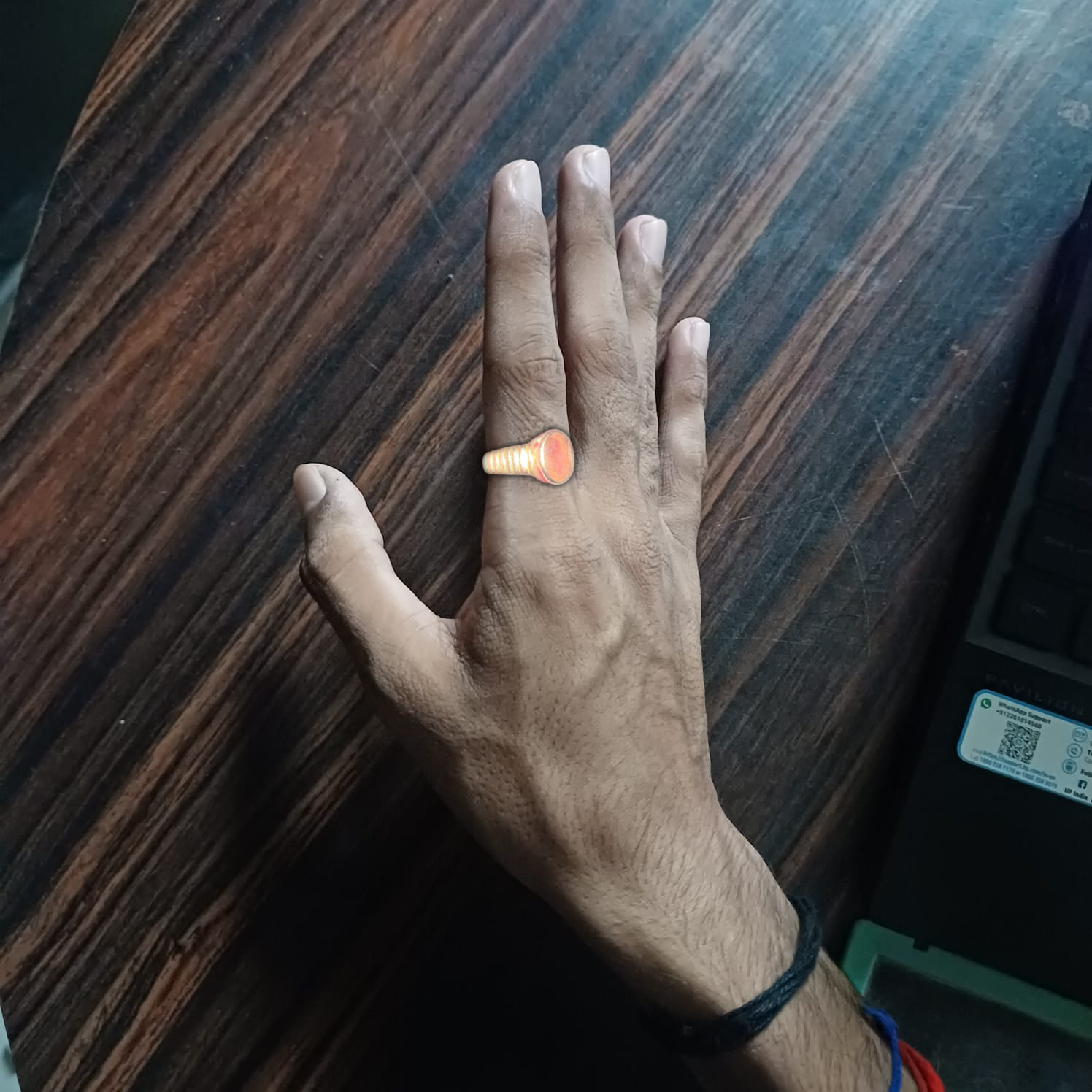}
        \caption{Proposed Approach}
        \label{fig:ours}
    \end{subfigure}

    \caption{Representative single-output comparison among all models.}
    \label{fig:singleoutputcomp}
\end{figure*}

\clearpage
\renewcommand{\arraystretch}{1.3}
\begin{table*}[t!]
\centering
\caption{Comparison of Jewellery Try-On Approaches}
\begin{tabular}{|p{3.2cm}|p{2.3cm}|p{2cm}|p{2.5cm}|p{2.5cm}|p{2.3cm}|}
\hline
\textbf{Approach} & \textbf{Realism} & \textbf{Computational Cost} & \textbf{Device Dependency} & \textbf{Fidelity to Input} & \textbf{Suitability for E-commerce} \\ \hline

Overlay-based approaches [1] & Very Low (2D static overlay without orientation) & Very Low & Works on all devices & Poor (ring not adapted to finger pose) & Limited (unnatural appearance) \\ \hline

AR-based solutions [2] & Medium (3D models with pose sync) & High (requires AR rendering) & Needs AR-enabled/high-end devices & Medium (pose noise causes misalignment) & Moderate (device accessibility issue) \\ \hline

Generative image synthesis [3] & Medium-High (global image regeneration) & Very High (GPU intensive) & Requires high-performance GPUs & Low (unwanted alterations in unmasked regions like the hand) & Limited (slow and inconsistent for users) \\ \hline

Diffusion-based inpainting [4] & High (localized edits) & High (GPU intensive) & Requires GPU-enabled systems & Medium (cannot perfectly reproduce reference ring) & Limited (computationally restrictive) \\ \hline

Proposed Approach & High (pose and orientation alignment, scaling preserved) & Medium (hybrid ML pipeline) & Runs on consumer smartphones & High (ring reproduced exactly with correct scaling) & High (realistic, efficient, deployable) \\ \hline

\end{tabular}
\end{table*}

\renewcommand{\arraystretch}{1.3}
\begin{table*}[t!]
\centering
\caption{Time comparison of all Approaches}
\begin{tabular}{|p{3.2cm}|p{2.3cm}|p{2cm}|p{2.5cm}|p{2.5cm}|p{2.3cm}|}
\hline
\textbf{Approach} & \textbf{Overlay Based methods} & \textbf{Candere by Kalyan} & \textbf{ OpenAI (ChatGPT)} & \textbf{Google AI Studio (banana pro)} & \textbf{Proposed Approach} \\ \hline

\textbf{Time Taken} & 3.32 sec & 40.02 sec & 2 min 23 sec & 38.64 sec & 34.42 sec \\ \hline

\end{tabular}
\end{table*}

As seen in Fig. 3(c) Google AI Studio (banana pro) keeps the background same, and the hand also remains consistent, same as the input. The ring is also getting generated but is not actually the same as the reference object, shown in Fig. 3(a), Input Ring, in our case. It tries to recreate the object based on the reference object, which results in a different-looking ring on the hand. This approach does not guarantee the exact generation of the reference object. This isn't acceptable in Try On Projects, in which both the input object and body part must remain as they are. It also requires quite an accurate prompt to generate even this type of accuracy, which is indeed not intended in a try-on use case. Also, it requires GPUs and high-end resources to generate output like this.

As seen in Fig. 3(d) OpenAI (ChatGPT) is regenerating the entire image based on both reference images of the hands and ring; hence, it does not guarantee background consistency. Google AI Studio uses the same hand image without trying to recreate it and only recreating the ring, but ChatGPT actually tries to recreate the entire image, and that results in a totally different output.

As seen in Fig. 3(e) Candere by Kalyan actually uses augmented reality to give this type of output. Now it actually looks unrealistic because size is not always accurate, and using the augmented reality approach, the output looks like a ring is floating on the finger, not attached to it. Also, the ring is not rotated properly with respect to finger rotation, which makes it more unrealistic.

As seen in Fig. 3(f) Proposed System we are not recreating the ring, not the hand. We actually calculate the size, rotation, and angle of the finger, and based on that, we also change the ring image rotation, size, and angle and put it on the finger and also apply the Gaussian blur to make it more realistic with lighting on the image. Compared to Google AI Studio and ChatGPT, we are not recreating the ring. Instead, we are actually using the same photos of hand and ring, which results in an unchanged hand and ring both. Also compared with canderers, we are calculating the size and applying Gaussian blur, which is attached with a finger and also relates to the lighting.

\section{Conclusion}

In conclusion, the proposed virtual jewelry try-on system provides a comprehensive, efficient, and easily available solution to the limitations of the try-on technologies which are currently in the market. The proposed system combines the strengths of both vision and geometric modeling. Unlike existing methods, which heavily rely on AR hardware or computationally demanding generative and diffusion models that are highly dependent on GPUs, this system adopts a lightweight approach that ensures both realism and accessibility on standard mobile devices. By combining MediaPipe [5] for precise hand landmark detection and YOLOv8 [6] for accurate ring detection and precise ring localization, the system achieves correct placement, alignment, and scaling of the jewelry according to the natural geometry of the user’s finger. The geometric transformation ensures that the ring’s orientation correctly fits to the user’s finger in a visually authentic manner, while alpha blending techniques enable the ring to match with the image synthesis and make it look realistic. The proposed system not only removes the need for GPUs and high-end AR-enabled devices, but it also makes sure that, while trying on the user's hand, the jewels both remain totally similar. Ultimately, this system shows how geometry-based solutions can deliver accurate and computationally efficient solutions that redefine virtual jewelry visualization for the modern digital marketplace.

\end{document}